\documentclass[journal]{IEEEtran}

\usepackage{times}
\usepackage{parskip}
\usepackage{subfig}
\usepackage{psfrag}
\usepackage{wrapfig}
\usepackage[latin1]{inputenc}
\usepackage{amssymb}
\usepackage{amsthm}
\usepackage{epsfig}
\usepackage{graphicx,times} 
\usepackage{amsfonts,amsmath,amssymb,array}
\usepackage{tabularx}   
\usepackage{colortbl}
\usepackage{longtable}
\usepackage{booktabs}
\usepackage{url}
\usepackage{icomma}
\usepackage{pstricks}
\usepackage{longtable}
\usepackage{multirow}
\usepackage{color}
\usepackage{graphicx}
\usepackage{cite}
\usepackage{picinpar}
\usepackage{amsmath}
\usepackage{amssymb}
\usepackage{mathtools}
\usepackage{url}
\usepackage{flushend}
\usepackage[latin1]{inputenc}
\usepackage{colortbl}
\usepackage{soul}
\usepackage{multirow}
\usepackage{pifont}
\usepackage{color}
\usepackage{alltt}
\usepackage{breakurl}
\usepackage{epstopdf}
\usepackage{pbox}
\usepackage{bbm}

\newtheorem{definition}{Definition}
\newtheorem{assumption}{Assumption}

\newtheorem{theorem}{Theorem}

\begin{document}

\title{Real Time Self-Tuning Adaptive Controllers on Temperature Control Loops using Event-based Game Theory}

\author{
\IEEEauthorblockN{Steve Yuwono\IEEEauthorrefmark{1}, Muhammad Uzair Rana\IEEEauthorrefmark{1}, Dorothea Schwung\IEEEauthorrefmark{3}, Andreas Schwung\IEEEauthorrefmark{1}}
\\ \IEEEauthorblockA{\IEEEauthorrefmark{1}South Westphalia University of Applied Sciences, Soest, Germany
    \\\{yuwono.steve, rana.muhammaduzair, schwung.andreas\}@fh-swf.de}
\\ \IEEEauthorblockA{\IEEEauthorrefmark{3} Hochschule D{\"u}sseldorf University of Applied Sciences, D{\"u}sseldorf, Germany
\\ dorothea.schwung@hs-duesseldorf.de}
}

\maketitle
\thispagestyle{empty}
\pagestyle{empty}

\begin{abstract}
This paper presents a novel method for enhancing the adaptability of Proportional-Integral-Derivative (PID) controllers in industrial systems using event-based dynamic game theory, which enables the PID controllers to self-learn, optimize, and fine-tune themselves. In contrast to conventional self-learning approaches, our proposed framework offers an event-driven control strategy and game-theoretic learning algorithms. The players collaborate with the PID controllers to dynamically adjust their gains in response to set point changes and disturbances. We provide a theoretical analysis showing sound convergence guarantees for the game given suitable stability ranges of the PID controlled loop. We further introduce an automatic boundary detection mechanism, which helps the players to find an optimal initialization of action spaces and significantly reduces the exploration time. The efficacy of this novel methodology is validated through its implementation in the temperature control loop of a printing press machine. Eventually, the outcomes of the proposed intelligent self-tuning PID controllers are highly promising, particularly in terms of reducing overshoot and settling time.
\end{abstract}

\def\abstractname{Note to Practitioners}
\begin{abstract}
Various industrial applications are controlled by Proportional-Integral-Derivative (PID) controllers. In fact, PID controllers are the go-to approach due to their ease of use and the potential to be adjusted by experienced control engineers. However, this adjustment process is notoriously time consuming and often results in suboptimal performance. This work proposes a generally applicable approach for automatic tuning of PID control parameters using a self-learning algorithm based on event-based state-based potential games. We discuss the design of the controller as well as certain convergence guarantees which help to maintain a stable control behavior during the controller tuning. We showcase the applicability and effectiveness of the approach on a temperature control loop for paper printing machines operated in different set points and under varying disturbances with highly promising results.
\end{abstract}

\begin{IEEEkeywords}
Self-tuning PID controller, game theory, intelligent control, artificial intelligence, temperature control
\end{IEEEkeywords}

\section{Introduction}\label{sec:intro}

Temperature control has been a fundamental and critical aspect of numerous industrial processes since it plays a pivotal role in ensuring operational safety, product quality, and process efficiency. Notably, industries such as chemical manufacturing, pharmaceuticals, and food production heavily rely on the precise regulation of temperature. In this context, it is imperative to avoid extreme temperature fluctuations. This principle holds in the broader manufacturing sector~\cite{Patrick2021}, where the prevention of machinery overheating is vital. This prevention facilitates uninterrupted machine operation, thereby boosting overall production efficiency. Furthermore, the precise control of temperature leads to a significant influence on energy efficiency and sustainability, which makes it a central concern for industries committed to reducing their environmental impact.

In numerous applications, the predominant and highly efficient method for temperature control within a system involves the utilization of Proportional-Integral-Derivative (PID) controllers~\cite{Ogata2010, Shein2012, Hamid2009, Mugisha2015}. A PID controller constitutes a feedback control system that modulates the control input of a process by considering the weighted combination of its proportional, integral, and derivative components to maintain the desired setpoint~\cite{Ogata2010}. Upon an extensive examination of the limitations and challenges in traditional PID control~\cite{Pandey2020, Su2020}, it is evident that traditional PID controllers face difficulties when confronted with dynamic and complex temperature control systems. The main contributing factors to these limitations are an inability to handle non-linearity, lack of robustness, slow response to alterations in setpoints, challenges in the context of multivariable systems, and the necessity for manual tuning.

One of the possible solutions involves the integration of intelligence by developing self-tuning PID controllers. The concept of self-tuning PID controllers lies in their ability to dynamically adjust to evolving system dynamics. Hence, this approach introduces adaptability and effectively addresses the above-mentioned limitations. There are several tuning methods for PID controllers, including heuristic approaches and offline methods like the Ziegler-Nichols Method~\cite{Ziegler1942} and Cohen-Coon Method~\cite{Joseph2018}. Self-tuning methods have been developed based on fuzzy-systems~\cite{Dehghani2015} or reinforcement learning (RL)~\cite{Dogru2022}. However, these methods either require expert knowledge or long and complex learning, presenting difficulties in practical applications. Hence, this study introduces an alternative and novel self-tuning method that leverages real-time data and self-learning techniques.

One of the effective self-learning approaches is the application of game theory within dynamic games, which has demonstrated success in various control system applications, as shown in~\cite{Schwung2023, Yuwono2022, Yuwono2023b}. Dynamic game theory, particularly state-based potential games (SbPG)~\cite{Schwung2022}, offers several advantages which enable the coordination of multiple control variables to work cooperatively in maximizing the global objectives of systems. Moreover, it proves to be a more practical choice for real-world industrial settings where neural networks are not required. Additionally, it provides an accurate and reliable framework for decision-making in dynamic and uncertain environments.

In this research, we enhance the game theory-based approach by introducing an event-based game theory as a self-tuning mechanism for PID controllers. Within this framework, players are not mandatory to update their actions at every time step but rather update their actions only when an event is triggered. Events can be initiated in response to system disturbances, alterations in setpoints, and other relevant factors. Our proposed approach employs a model-free strategy, which eliminates the need for a detailed mathematical model of the system.


Our study makes the following contributions:
\begin{itemize}
    \item We propose a novel intelligent self-tuning PID controller based on game theory. We present the design of the game as well as suitable self-learning algorithms for automatic tuning. 
    \item We propose event-based SbPG as an extension to classical SbPG, which prevents unnecessary action changes, enabling control parameters updates only when they are necessary.
    \item We conduct a theoretical analysis of the resulting game-based controller resulting in sound convergence guarantees of the tuning process.
    \item We showcase the practical application of the proposed event-based game theory approach to temperature control for a printing press machine, which reveals noteworthy and significant outcomes.
\end{itemize}

The structure of this paper is organized into six sections. Section~\ref{sec:review} explores prior works, methodologies, and advancements in PID controllers and game theory for industrial automation. In Section~\ref{sec:sbpg}, the core of our research is presented, which outlines the proposed approach involving event-based game theory in dynamic games. Section~\ref{sec:exp} focuses on incorporating this novel method into PID controllers, which results in the development of intelligent self-tuning PID controllers. Subsequently, Section~\ref{sec:test} presents the sample industrial-related practical experiment, along with the analysis and discussion of the results. Finally, in Section~\ref{sec:conclusion}, we conclude and explore extensions for future research.

\section{Literature Review}\label{sec:review}
In this section, we discuss the state-of-the-art and related works on PID controllers for industrial automation and game theory for self-optimizing manufacturing systems.

\subsection{PID Controllers for Industrial Automation}\label{sec:rev_1}

PID controllers represent a class of control systems widely used in industrial automation~\cite{Ogata2010}. Their primary function is to regulate and maintain desired process variables within predefined boundaries. The PID controller contains three main components, namely, proportional, integral, and derivative~\cite{Ogata2010}. In mathematical terms, a PID controller can be expressed as:
\begin{equation}\label{eq:pid}
u(t)=K_p e(t) + K_i \int_0^t e(t)dt + K_d \frac{de(t)}{dt},    
\end{equation}
where $u(t)$ denotes the controller output, $e(t)$ refers to the error at time $t$, and $K_p, K_i, K_d$ are the proportional, integral, and derivative gains.

PID controllers have been applied in many industrial sectors and offer several advantages in industrial automation. These advantages include their simplicity, ability to ensure system stability, and cost-effectiveness. One of the widespread industrial applications of PID controllers is in closed-loop temperature control, as studied in~\cite{Shein2012, Hamid2009, Mugisha2015}. Although PID controllers offer several advantages, PID controllers have certain challenges and limitations~\cite{Pandey2020, Su2020}, including tuning complexity, non-linearity, and vulnerability to disturbances. 

Standard methods for tuning PID controllers rely on experimental adjustments based on the system response. The Ziegler-Nichols~\cite{Ziegler1942} method and subsequent developments~\cite{Joseph2018}, are based on certain signal characteristics (e.g. decay ratio, settling time) and offer practical approaches for setting parameters. However, these methods are designed to steer the system into an oscillatory state, which can be problematic in sensitive environments.

Various approaches have been developed to cope with these challenges. Specifically, \cite{Pandey2020} propose PID auto-tuning based on ON/OFF control, \cite{Su2020} use the Levenberg-Marquart algorithm while \cite{Moura2020} employ neuro-fuzzy models. Further approaches based on fuzzy-systems have been reported~\cite{Dehghani2015, Flores2018}. All of these algorithms are designed for specific system structures and require different levels of knowledge about the system.

Another line of research proposes the use of RL~\cite{sutton2018reinforcement} for PID parameter tuning. Specifically, \cite{Dogru2022} uses RL for tuning with constraints integrating a system identification step with subsequent fine-tuning of the parameters. RL-based adaptive PID-control is introduced in~\cite{SHUPRAJHAA2022109450} specifically tailored for nonlinear unstable processes. The work~\cite{ces.2021.15} specifically concentrates on stability preservation when using PID parameter tuning. A novel entropy maximization based RL-algorithm for automatic tuning has been proposed in~\cite{10156246}. A real-world application of RL-based PID tuning is presented in~\cite{LAWRENCE2022105046} discussing various implementation aspects on standard hardware.

However, RL algorithms are notoriously difficult to train, requiring a huge number of system interactions with varying system characteristics which render their real-time application challenging. Furthermore, these algorithms typically learn continuously in the sense that learning updates have fixed sampling times. However, this is unreasonable, as parameter tuning is only required if the system response is not sufficient and the system is sufficiently excited. Also, RL does not allow for a coordinated training of the control parameter. 

In this work, we bridge this gap by proposing event-based game theory based PID tuning which results in simple, easy to implement learning updates based on performance based event triggers. The approach allows for coordinated learning of PID parameters under sound convergence guarantees.

\subsection{Game Theory for Self-Optimization and Control}\label{sec:rev_2}

Game theory is a mathematical framework that deals with strategic interactions between multiple decision-makers~\cite{Owen2013, Bauso2016}. Game theory has found various applications in control systems, ranging from $H_{\infty}$-control~\cite{Limebeer1992} to cooperative control~\cite{Fele2017}, see also the survey in~\cite{Marden2018ab}. However, the application of game theory to PID-controller tuning has not been discussed.

Furthermore, a limited number of game theory-based optimization approaches have been presented recently. Stackelberg games are employed for demand response schemes in smart manufacturing~\cite{Lee2018}, collaborations in third-party remanufacturing~\cite{Wang2019} and distributed control~\cite{Yuwono2024b}. Combining mixed-integer optimization and noncooperative games is presented in~\cite{Liu2019}. Cooperative games are introduced for cloud manufacturing platforms~\cite{Chao2018}. However, all of these approaches cover the planning level of optimization while we concentrate on the field level control.

A particularly appealing game structure are potential games~\cite{Monderer1996} which recently found applications in control including smart energy systems~\cite{Liang2018}, wind turbine control~\cite{Marden2013}, resource allocation~\cite{La2016} and control of communication systems~\cite{Yamamoto2015}. However, potential games (PGs) lack state information limiting their applicability in dynamic systems. Consequently, SbPG~\cite{Marden2012} and dynamic PG~\cite{Zazo2016} have been introduced recently and applied to communication systems~\cite{Zazo2016}, modular production systems~\cite{Schwung2022} and PLC-informed optimization~\cite{Schwung2023}. Incorporation model information within SbPG-based distributed control has been presented in~\cite{Yuwono2023b}. All of these approaches apply SbPG as a framework for iterative learning from scratch similarly to RL optimization. Contrary, we use PID controller as the backbone controller in which we incorporate SbPG-based learning for the autotuning of PID controllers.

\section{Event-based Game Theory in Dynamic Games}\label{sec:sbpg}

In this section, we introduce event-based dynamic games. This method builds upon established self-learning algorithms, such as SbPGs~\cite{Schwung2022}, with the main distinction that players do not require constant adjustment of their actions. Rather, they only respond when specific events take place.

\subsection{State-based Potential Games}\label{sec:sbpg_1}

PGs~\cite{Monderer1996} represent strategic games where players aim to optimize their local utility $U_i$ with respect to a global objective function, known as the potential function $\phi$. Given that PGs specify a set of \textit{N} players the strategic-form game in PGs is constructed as $\Gamma(\mathcal{N}, \mathcal{A}, \{U_i\}, \phi)$, where $\mathcal{A}$ represents the set of individual actions $a_i$. Due to the specific game structure, PGs converge to Nash equilibrium~\cite{Marden2012}, where no player can improve their outcome by unilaterally changing their strategy. 

A limiting factor in PGs is that the system behavior is considered to be static, which is not true in various applications. Hence, the concept of PGs has been extended to SbPGs~\cite{Schwung2022}, which are particularly well-suited for self-control and optimization. In SbPGs, the game is constructed as $\Gamma(\mathcal{N}, \mathcal{A}, \{U_i\}, {S}, P, \phi)$ with the set of states \textit{S} and the state transition process \textit{P}. A game qualifies as an SbPG if a potential function $\phi:  a \times  {S} \rightarrow \mathcal{R}^{c_i}$ can be derived satisfying the following conditions for each pairing of state-action $[a,s]\in  \mathcal{A}\times {S}$:
\begin{align}\label{eq:potcondsbpg}
U_i(a_i,s) - U_i({a}^{\prime}_i, {a}_{-i},s) = \phi(a_i,s) - \phi({a}^{\prime}_i, {a}_{-i},s)
\end{align}
and
\begin{align}\label{eq:condsbpg}
\phi(a_i,s^{\prime}) \geq \phi(a_i,s).
\end{align}
These conditions hold for any state $s^{\prime}$ in $P(a,s)$ and $\mathcal{R}^{c_i}$ represents continuous actions.

The above definition of SbPG allows to maximize the overall utility of the system 
\begin{align}\label{eq:optproblem}
\max_{{a}\in \mathcal{A}} \phi({a},{S}).
\end{align}
by collectively maximizing the local utilities $U_i(a_i, s)$ of the players. Convergence guarantees to a Nash equilibrium remain valid for SbPG~\cite{Marden2012,Zazo2016} under suitably designed utility functions. Criteria for the design of utility functions are reported in~\cite{Zazo2016,Schwung2022}.

\subsection{Optimization Algorithms for SbPG}\label{sec:learn_1}

\begin{figure}[t]
	\centering
	\includegraphics[width=0.85\linewidth,keepaspectratio]{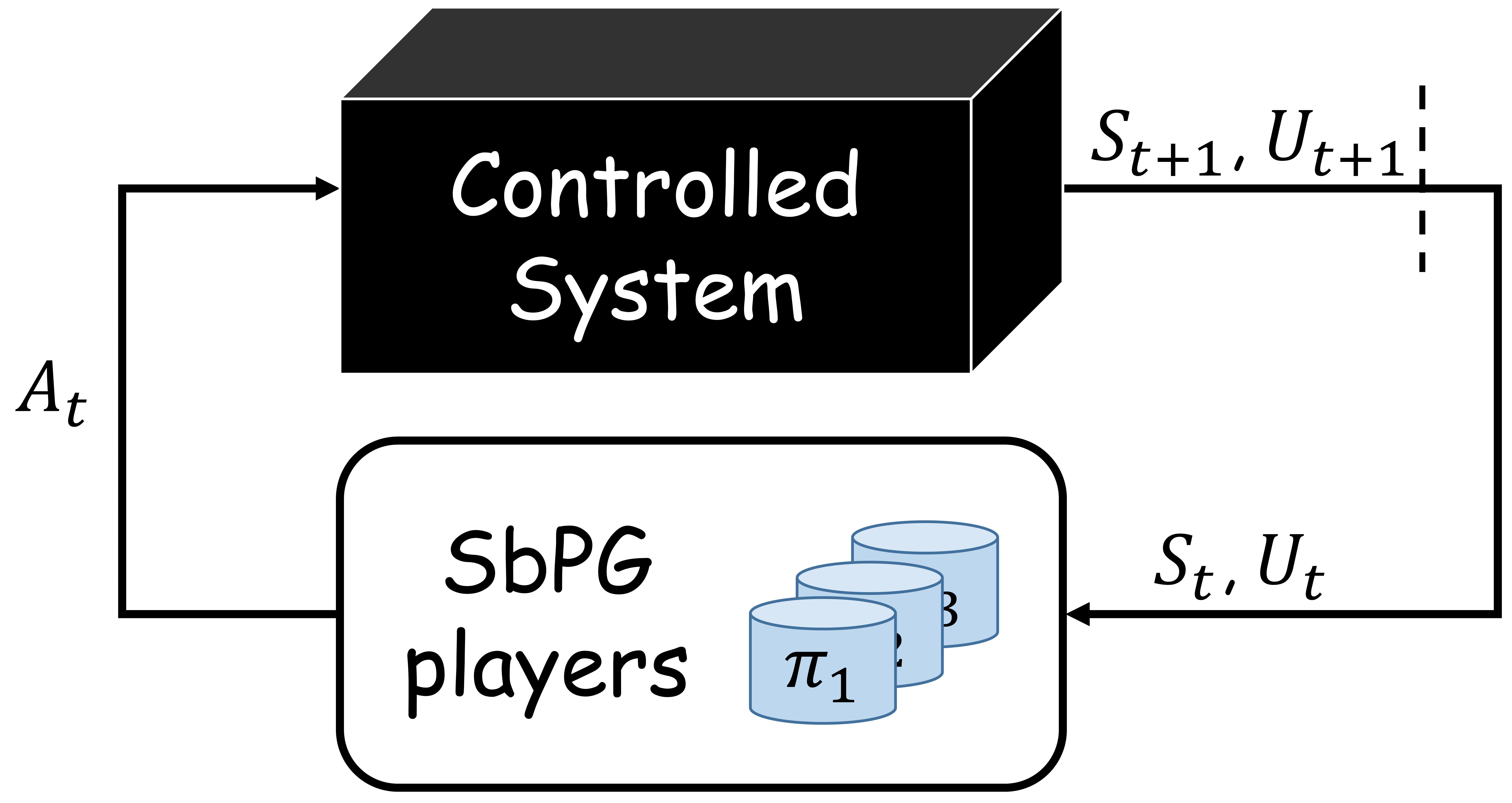}
	\caption{A schematic diagram of SbPGs.}
	\label{fig:sbpg}
\end{figure}

The optimization process of SbPGs typically follows a distributed learning paradigm. Specifically, the optimization takes place while interacting with the environment in a model-free optimization scheme similar to RL~\cite{sutton2018reinforcement} as illustrated in Fig.~\ref{fig:sbpg}.
While the application of RL algorithms for SbPG is possible, computationally simpler learning algorithms with higher performance have been proposed including best response~\cite{Schwung2022} and gradient based learning~\cite{Yuwono2024}.

\begin{figure}[t]
	\centering
	\includegraphics[width=0.65\linewidth,keepaspectratio]{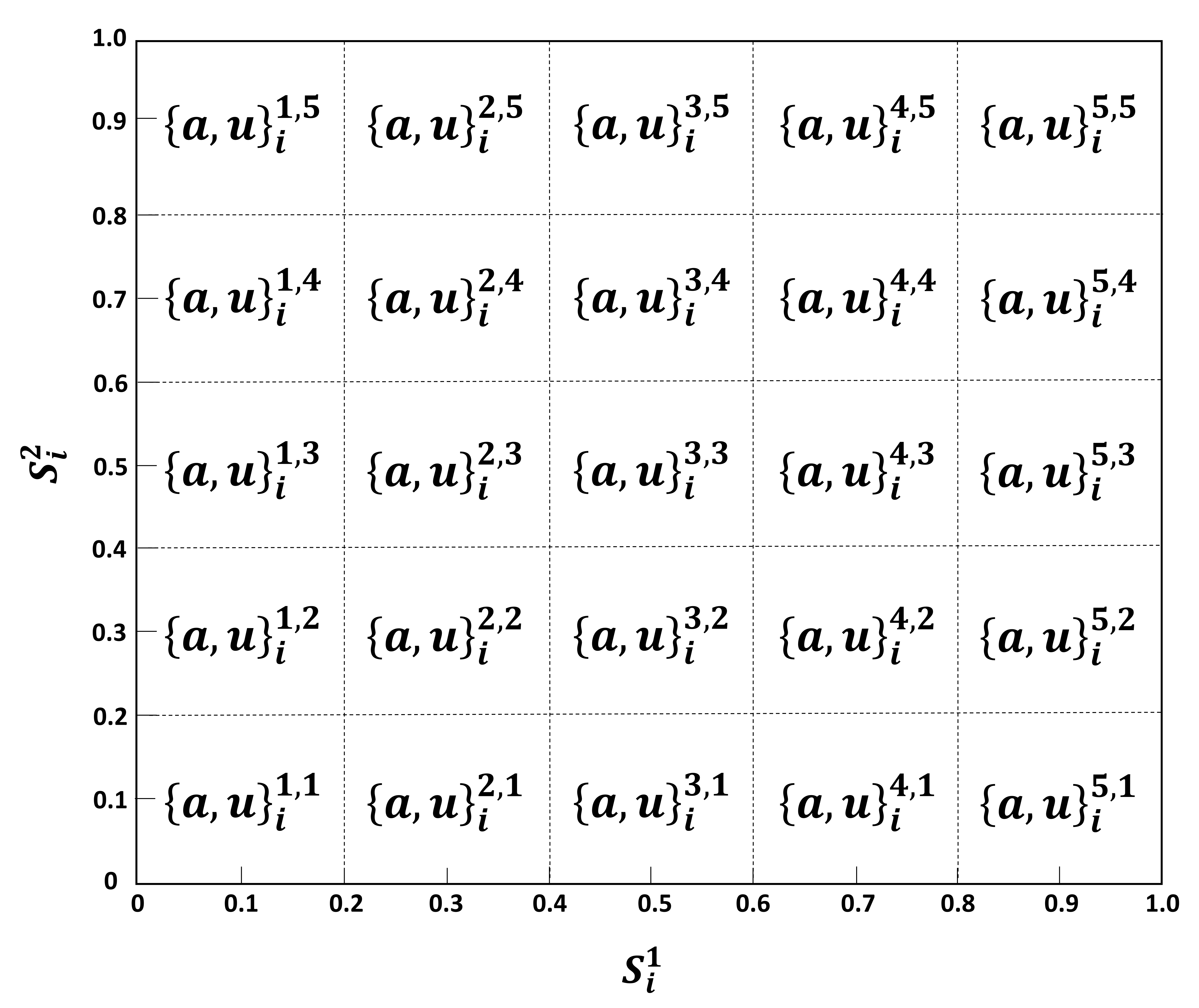}
	\caption{An illustrative 2D performance map of state-action-utility combination within SbPGs.}
	\label{fig:maps}
\end{figure}

Both learning schemes have in common that each player $i$ has individual performance maps, which store the optimal actions and their corresponding utility values across various state combinations, as illustrated in Fig.~\ref{fig:maps}. These performance maps are continually updated and optimized during the training phase, which involves the exploration of different actions within state combinations. The updates can be done either in a best response manner by storing optimal utility values or by approximating gradients within the optimization region. While the former algorithm is computationally cheaper, the latter offers faster optimization.

Based on the obtained performance maps, a player $i$ selects an action \textit{$a_{i,t+1}$} through global or local interpolation techniques applied to these performance maps. The global interpolation process is as follows:
\begin{equation} 
    w_i^{\overline{s^0s^q}} = \dfrac{1}{{(d_i^{s^0s^q})}^2 + \gamma_{map}},
\end{equation}
\begin{equation} 
    a_{i} = \sum_{q} \dfrac{w_i^{\overline{s^0s^q}}}{\sum_{q}w_i^{s^0s^q}} \cdot a_i^q,
\end{equation}
where $s^0$ is the current state, $s^q$ refers to the state of the $q$-th support vector, $d_i^{s^0s^q}$ represents the absolute distance between $s^0$ and $s^q$, $w_i^{\overline{s^0s^m}}$ is the computed weight, and $\gamma_{map}$ is a smoothing parameter.

We highlight the point that in prior research, players are required to continually adjust their actions in every time step $a_{i,t+1}$ based on the current states $s_{i,t}$. However, for PID controller tuning, this may not be reasonable, as a continuous update of the control parameters is neither required nor desired, requiring event-based SbPG as formalized next.

\subsection{Event-based SbPGs}\label{sec:sbpg_2}

The primary goal in introducing an event-based method to SbPGs (Eb-SbPGs) is to avoid the necessity for players to alternate their actions in every time step and restrict the alternation of actions to instances when an event is triggered. The utilization of Eb-SbPGs is particularly significant in the domain of control systems applications. In any control system, a collection of uncertainties and disturbances exist, which can potentially lead to system performance degradation and unfavourable outcomes. Furthermore, it typically requires a certain amount of time for a controlled system to achieve a steady state following disturbances, uncertainties, or adjustments to set points. The mentioned conditions can serve as triggers for new events, and the Eb-SbPGs framework accommodates players to promptly respond and adapt their actions in response to these triggered events. Following~\cite{6425820}, trigger times for events can be defined as 
\begin{align}
    t_0\!=\!t^*_0\!=\!0, \ & t_{k+1} = \inf\{t \in \mathbb{R}\ |\ t > t^*_k \wedge G_e(t)\!=\!0\},\\
    & t^*_{k+1} = \inf\{t \in \mathbb{R}\ |\ t > t_{k+1} \wedge G_r(t)\!=\!0\},
\end{align}
where $G_e(t)$ is a suitably defined trigger function and $G_r(t)$ indicates a reset function indicating that the system has regained its desired performance.

Based on the above elaborations, we now extend the SbPG framework to Eb-SbPGs as follows:
\begin{definition}\label{def:eb_sbpg}
    A game $\Gamma(\mathcal{N}, \mathcal{A}, \{U_i\}, \mathcal{S}, P, G_e, G_r, \phi_F)$ is an event-based SbPG, if it satisfies the following conditions:
    \begin{align}\label{eq:potcondsbpg_follower}
    U_i(a_i,s) - U_f({a}^{\prime}_i, {a}_{-i},s) = \phi(a,s) - \phi({a}^{\prime}_i, {a}_{-i},s),
    \end{align}
    and
    \begin{align}\label{eq:condsbpg_follower}
    \phi(a_i,s^{\prime}) \geq \phi(a_i,s),
    \end{align}
    for any state $s^{\prime}$ in $P(a,s)$ and $\forall t\in [t_{k}, t^*_{k}]$.
\end{definition}
Hence, in contrast to SbPG, the update mechanism only takes place once an event is triggered until certain performance conditions are met.

\section{Intelligent Self-Tuning PID Controllers}\label{sec:exp}

In this section, we employ the Eb-SbPG framework for intelligent self-tuning PID controllers using the Eb-SBPG framework. 

\subsection{Self-Tuning PID Controllers using Eb-SbPG}\label{sec:exp_2}

As stated before, default PID control strategies reveal several limitations, as it requires manual setup of the initial values for $K_P$ and $K_I$, followed by potentially extensive trial and error procedures to identify the most suitable control parameters. More worse, we could not guarantee that the manually chosen parameters represent an optimized solution. 

\begin{figure}[t]
	\centering
	\includegraphics[width=1.00\linewidth,keepaspectratio]{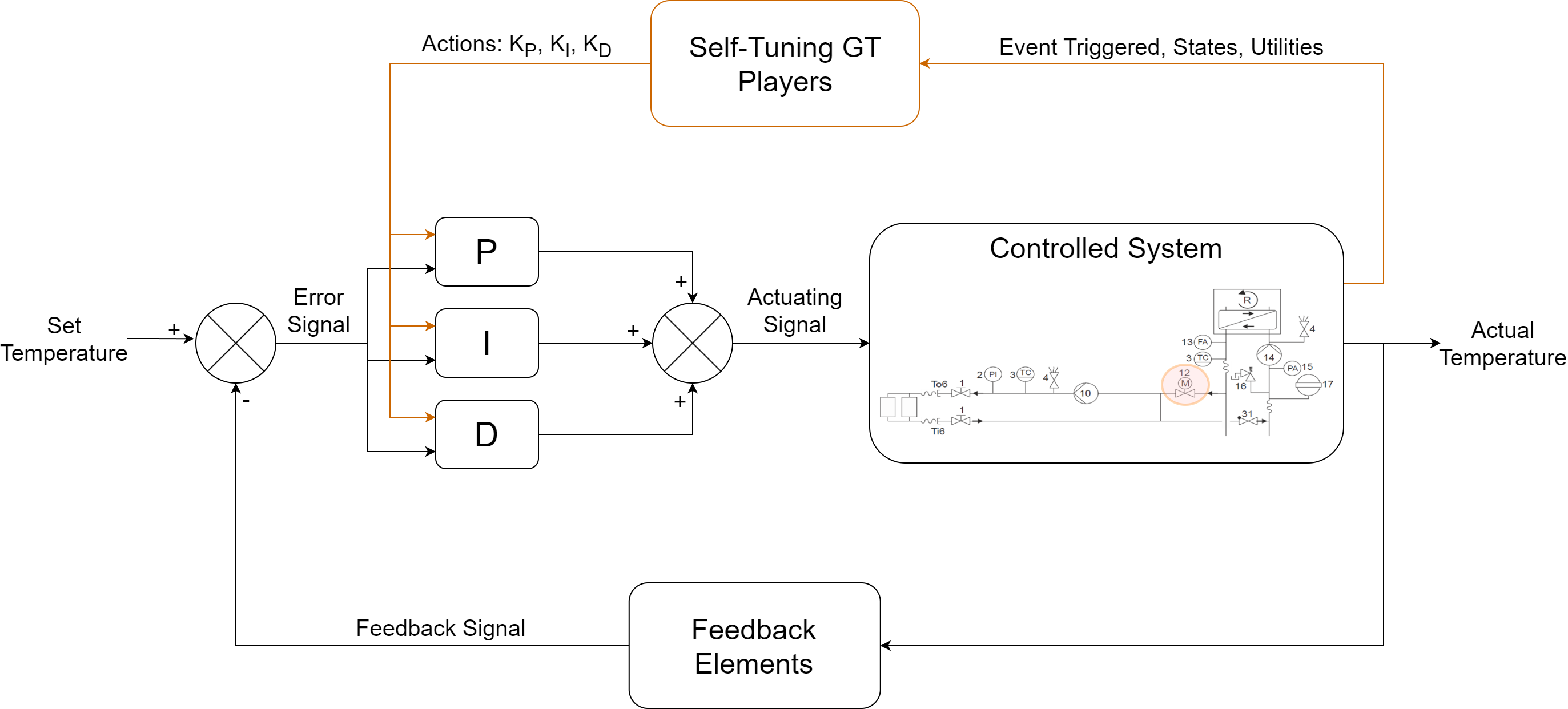}
	\caption{An illustration demonstrating the integration of an Eb-SbPG as a self-tuning PID controller within a closed-loop control system.}
	\label{fig:closedloop_gt}
\end{figure}

In response to these limitations, we propose to use the Eb-SbPG framework as a scheme for intelligent self-tuning PID controllers, as shown in Fig.~\ref{fig:closedloop_gt}. To this end, we have to specify various components of the Eb-SbPG including its state and action space, the definition of the utility functions of each player as well as the trigger function $G_e(t)$ and reset function $G_r(t)$. In addition, as we will see later on, we have to specify a range of control parameter settings, for which the system is known to be stable. We note that the approach is model-free in the sense that we do not require the state transition function to be known. Also, we note that the potential function $\phi$ is not required for the optimization but can be derived from the individual utility functions~\cite{Zazo2016}. 

\subsubsection{Action Set}\label{sec:a_set}

As shown in Fig.~\ref{fig:closedloop_gt}, we define the players of the game as the control parameters $K_P$, $K_I$ and $K_D$. For each player, we define a continuous action set $\mathcal{K}_x = \{K_x \in \mathbb{R} \lvert K_{x,\min} \leq K_x \leq K_{x,\max}\}$ for $x\in {P,I,D}$ which indicate the individual control parameter ranges. Consequently, the resulting action set is $\mathcal{A} = \bigcup_{x\in \{P,I,D\}} \mathcal{K}_x$.

\subsubsection{State Set}\label{sec:s_set}

In contrast to the action set, the state set is highly application-dependent. However, common to all applications is the use of the controller deviation $e$ as state variable. Hence, we have
\begin{align}
    s_i^e = e = y_{set} - y
\end{align}
where $s_i^e \in \mathbb{R}^m$ is the state of the Eb-SbPG related to the controller deviation, $y \in \mathbb{R}^m$ is the system output of dimension $m$ and $y_{set} \in \mathbb{R}^m$ is the desired setpoint. Note that we might use the absolute value $\lvert e \rvert$ instead of the controller deviation $e$ if the sign of the deviation is not important to rate the control quality. The state can be augmented with additional internal and external variables as e.g. states representing measurable disturbances $s_i^d \in \mathbb{R}^r$ of dimension $r$.

\subsection{Trigger Function Design}\label{sec:trigger_func}

\begin{figure}[t]
	\centering
	\includegraphics[width=1.00\linewidth,keepaspectratio]{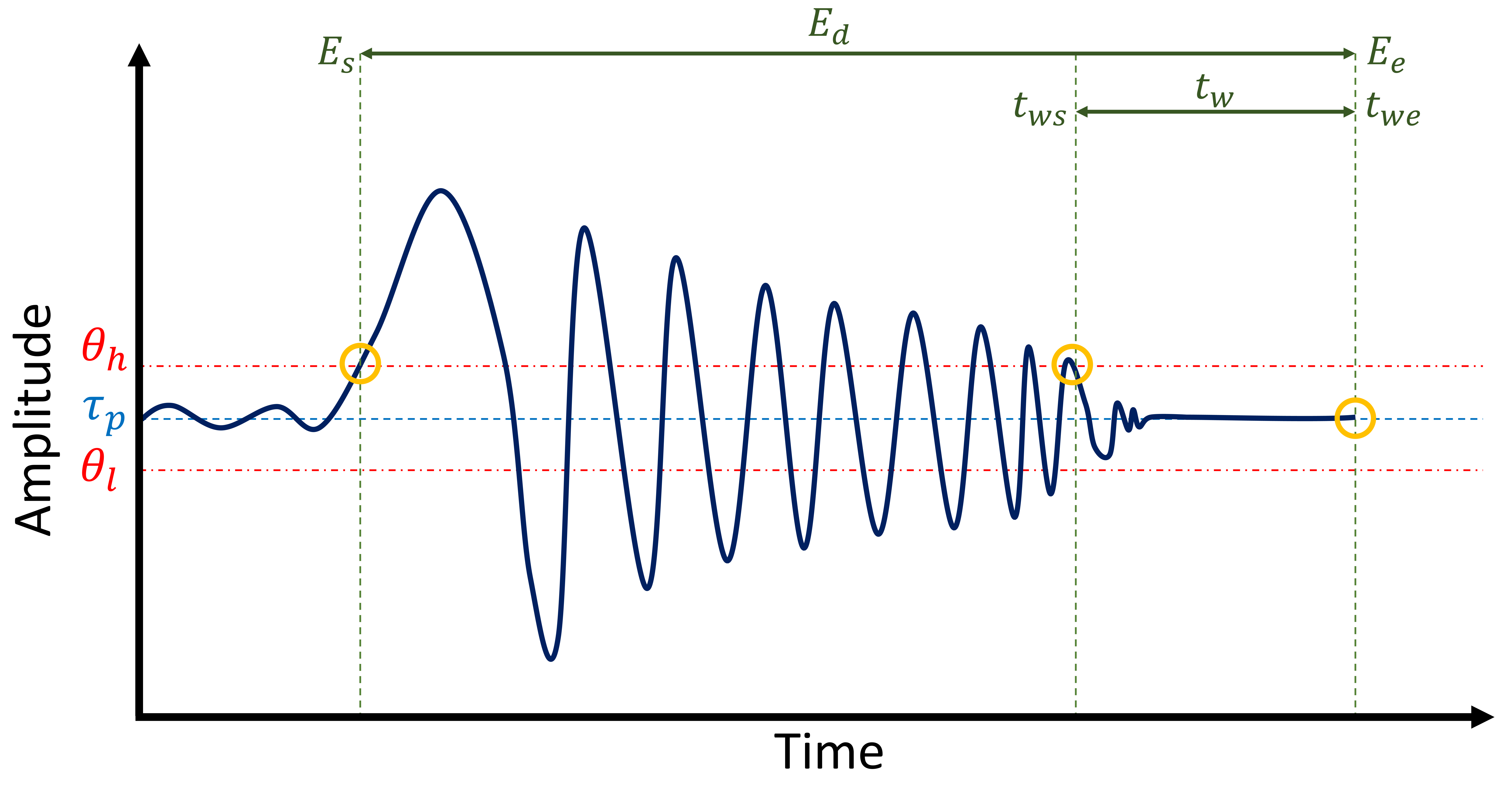}
	\caption{An illustration of an event trigger within the framework of Eb-SbPGs.}
	\label{fig:event}
\end{figure}

The purpose of the trigger functions $G_e$ and $G_r$ is twofold. First, it shall indicate a loss and a regain of performance of the controller under operation. Second, for tuning, a sufficient level of excitation is required, which shall be ensured. Fig.~\ref{fig:event} provides a visual representation of such a scenario, indicating a trigger event, where $\tau_p$ signifies the desired set point to be maintained, while $\theta_h$ and $\theta_l$ denote the upper and lower threshold boundaries respectively. These boundaries are considered hyperparameters and can be defined based on the sensitivity of the controlled system. If the actual amplitude surpasses either of these boundaries for any reason, it can be inferred that the control performance is no longer satisfactory while the loop has sufficient excitation, and an event is triggered (indicated by $G_e>0$). Consequently, when an event is triggered at $t_k$, the players receive a signal to adjust their actions $a_{i}$ in response to the current states $s_{i,t_k}$.
We note that Fig.~\ref{fig:event} illustrates a simple example using a fixed threshold. More advanced trigger function $G_e$ using adaptive thresholds or control envelopes can be easily integrated into the approach.

After the event is triggered, we have to define a reset trigger indicating the regain of the desired operation. To this end, we continuously monitor the control performance and identify instances when it remains within the predefined boundaries, where this moment is denoted by $t_{ws}$. If the amplitude remains within these boundaries for a specified duration $t_w$ serving as a hyperparameter, we can infer that the system has regained the performance. Consequently, both the detection mechanism and the associated event can be terminated, which are denoted by $t^*_{k}$ and $G_r=0$, respectively.

\subsection{Utility Function Design}\label{sec:exp_2_2}

The utility design is at the core of the SbPG-based PID tuning as it describes the control objective of the optimization approach. The design has two main objectives. First, we have to find suitable criteria to evaluate the control performance over the horizon $[t_{k}, t^*_{k}]$. Second, we have to assure that the control parameter remains within the correspondingly defined action set.

We propose a distributed learning scheme in the sense that both players maintain separate performance maps of actions and utilities. Hence, the utility functions of both players are different. We propose two variants of the utility function for self-tuning control. The first version is defined as:
\begin{align}\label{eq:util_centralized_1}
U_i \!=\! \alpha_{x} \frac{E_d(s_i^c)}{\|s_i^c\|_1\!+\!\|s_i^d\|_1\!+\!\gamma} \frac{1}{H_{m}(s_i^c)} + \alpha_{y} \frac{1}{H_{m}(s_i^c)}\!-\!B(a_i).
\end{align}
The second version is defined as 
\begin{align}\label{eq:util_centralized_2}
U_i \!=\! \alpha_{x} \frac{E_d(s_i^c)}{\|s_i^c\|_1\!+\!\|s_i^d\|_1\!+\!\gamma}  + \alpha_{y} \frac{1}{H_{m}(s_i^c)}\!-\!B(a_i).
\end{align}
In both utility fucntions, $E_{d}(s_i^c) = t^*_{k}-t_{k}$ denotes the settling time of the system, $H_{m}(s_i^c)$ denotes the maximum overshoot or undershoot during the event and $\alpha_{x}$ and $\alpha_{y}$ represent the auxiliary weight parameters. The above utility function encourages small over- and undershoots, low control deviations and short settling times which are common control criteria during parameter tuning.

As the above utility functions might steer the action set out of the given range, we incorporate a barrier function $B(a_i)$ designed to maintain the parameter within the range. To simplify the function design and to reduce the influence of the barrier function within the given action set, we propose the following function:
\begin{align}
    B(a_i) = \begin{cases}
        -\alpha_i (a_i\!-\!a_{i,\min}),& \text{if} \ \ a_i \!\leq\! a_{i,\min} \\
        0,& \text{if} \ \ a_{i,\min} \!<\! a_i \!<\! a_{i,\max} \\
        \alpha_i (a_i\!-\!a_{i,\max}),& \text{if} \ \ a_i \!\geq\! a_{i,\max}
    \end{cases}
\end{align}
where parameters $\alpha_i$ are constants whose values will be discussed below. At the end of an event, the players receive their respective utilities $U_i$ and proceed to optimize their policy. The selected actions are retained until the occurrence of another triggered event.

\subsection{Convergence and Stability}\label{sec:theory}

We now analyze the theoretical properties of the resulting self-tuning PID controller. Particularly, we focus on conditions on the stability of the closed loop system as well as the convergence properties of the Eb-SbPG. We start with assumptions before stating the main theorem.
\begin{assumption}\label{assump1}
    The action set $\mathcal{A}$ is chosen, such that the closed control loop is stable for all $a \in \mathcal{A}$.
\end{assumption}
Stable parameter sets can be derived using various methods including root locus and Lyapunov approaches depending on the properties of the controlled system. Alternatively, we can use a simple automated mechanism mimicking the Ziegler-Nichols method for identifying the potential stable region within the action space. This is achieved by implementing a grid search approach as outlined in~\cite{Bergstra2012} that takes place before initiating the training process for the Eb-SbPGs and allows to define the stable action set.
\begin{assumption}\label{assump2}
    The parameter $\alpha_i$ of the barrier function is chosen to be greater than the infimum of the approximated gradient of the utility function $\hat{U}_i$ with respect to the controller parameter, i.e.
    \begin{align}
        \alpha_i > \inf_{s,a} \left(\frac{\partial \hat{U}_i(s_i,a)}{\partial a_i} \right).
    \end{align}
\end{assumption}
This is a standard assumption on barrier function design and easy to fulfill with suitably large parameters $\alpha_i$.
\begin{theorem}
    Given Assumptions~\ref{assump1} and~\ref{assump2}, the Eb-SbPG, i.e. the control parameters, converge to a Nash equilibrium. Moreover, the resulting control loop is stable.
\end{theorem}
\begin{proof}
    We begin with the proof of the convergence of the controller parameter within the SbPG. To this end, according to Theorem in~\cite{Zazo2016}, we have to prove that the Hessian matrix of the game is symmetric, i.e. we have to prove that
	\begin{align}\label{eq:dpg1}
	\frac{{\partial}^2U_i(s_i,a)}{{\partial}a_j{\partial}s_m}=\frac{{\partial}^2U_j(s_j,a)}{{\partial}a_i{\partial}s_n},
	\end{align}
	\begin{align}\label{eq:dpg2}
	\frac{{\partial}^2U_i(s_i,a)}{{\partial}s_n{\partial}s_m}=\frac{{\partial}^2U_j(s_j,a)}{{\partial}s_m{\partial}s_n},
	\end{align}
	\begin{align}\label{eq:dpg3}
	\frac{{\partial}^2U_i(s_i,a)}{{\partial}a_j{\partial}a_i}=\frac{{\partial}^2U_j(s_j,a)}{{\partial}a_i{\partial}a_j},
	\end{align}
	$\forall i,j \in \mathcal{N}$, $\forall m \in S^{A_i}$ and $\forall n \in S^{A_j}$.\\
    It is easy to see that Cond.~\eqref{eq:dpg1} and~\eqref{eq:dpg3} are both fulfilled, as both sides equate to zero given the utility function design. Moreover, as the state-dependent part of the utility is shared among the players, both sides of Cond.~\eqref{eq:dpg2} are equal. Hence, as the game constitutes a SbPG, the condition for convergence to a Nash-equilibrium is assured. However, even though this condition is approved, stability is not preserved as the Nash equilibrium potentially lies outside of the stable parameter set. On the other hand, according to Assumption~\ref{assump2}, the gradient of the barrier function is designed to be greater than the approximated gradient of the SbPG utility. Hence, the gradient of the corresponding barrier function will always keep the parameters update within their defined action set. As by Assumption~\ref{assump1} this set includes parameters settings with stable control loop only, this concludes the proof. 
\end{proof}
We remark that the above convergence proof of SbPG holds for various utility function designs including utilities with varying state-dependencies among the players, as long as Cond.~\eqref{eq:dpg2} is fulfilled.

\section{Application to Temperature Control Loops}\label{sec:test}

We implement this novel methodology within a closed-loop feedback control system, specifically in the context of temperature control within a printing press machine.

\subsection{Temperature Control Loop on Printing Press Machine}\label{sec:exp_1}

\begin{figure}[t]
	\centering
	\includegraphics[width=1.0\linewidth,keepaspectratio]{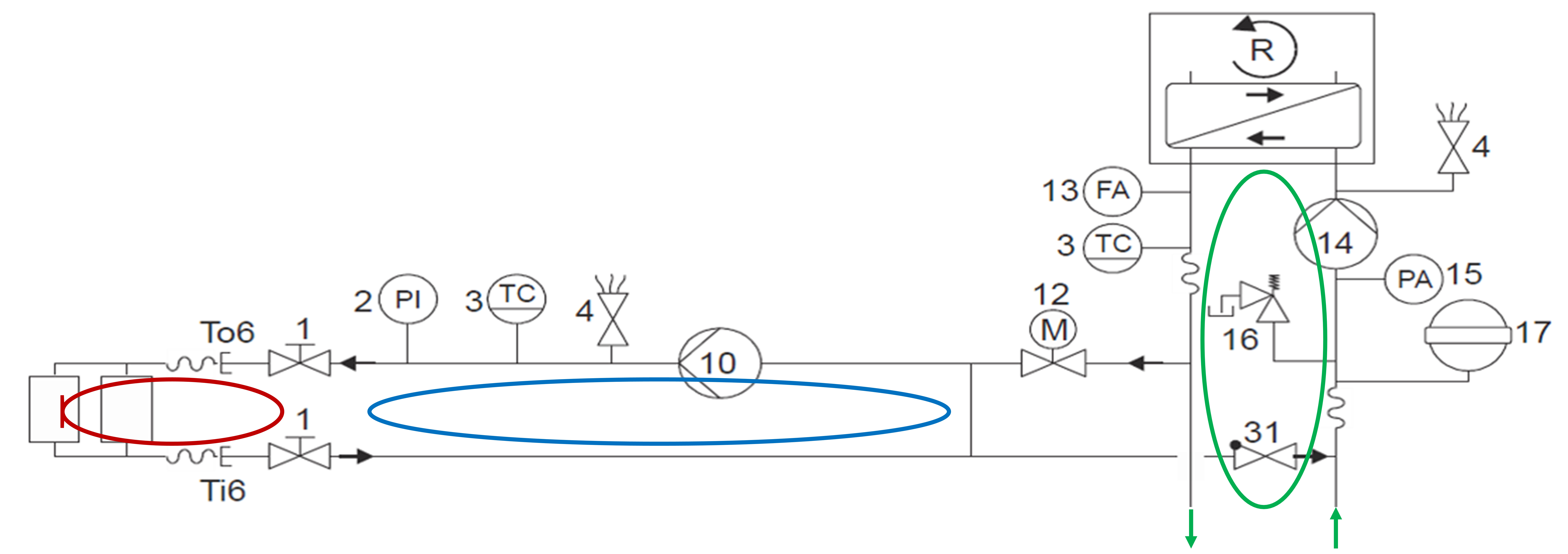}
	\caption{A piping and instrumentation diagram of the temperature control loop.}
	\label{fig:pid}
\end{figure}

The system is employed for regulating the temperature of the printing rollers in printing presses. In Fig.~\ref{fig:pid}, the temperature control circuit is highlighted in blue, the circulation circuit is in green, and the interface for connecting external loads is in red. To be noted, the circulation circuit is linked to multiple temperature control circuits within the system.

The optimal operation of these units requires keeping a consistent temperature $T_T$ by means of the control circuit. The heat $Q_L$ generated by the exothermic printing rollers is transferred to this temperature control circuit, which results in a temperature increase. Additionally, the ambient temperature $T_U$, the pressure differential $\Delta p$ between the temperature control circuit and the circulation circuit, and the volume $V_T$ of the temperature control circuit influence the process and serve as potential disturbances. The main objective of the controller is to maintain the temperature of the temperature control circuit $T_T$ at a constant level by supplying a specific volume of cooling water from the circulation circuit $T_Z$. The secondary objective is to keep $T_Z$ constant to avoid interference with other downstream temperature control circuits.

\subsubsection{Components, Hardware, and Software}\label{sec:exp_1_1}
Within the circulation circuit, the flow is maintained by a pump (see Component 14 in Fig.~\ref{fig:pid}) and the temperature is cooled through a compressor, in which the temperature and flow rate are both monitored using sensors (3, 13) at the outlet of the heat exchanger. Other components, including the rapid deaerator (4), the pressure switch (15), the safety valve (16), and the expansion vessel (17) are connected to the circulation circuit.

The temperature control circuit features a pump (10) to maintain a consistent flow rate as well as sensors (2, 3) for measuring water pressure and temperature. This pump operates discretely (on/off) ensuring a steady flow. Cold water is supplied from the circulation circuit to the temperature control circuit through a motor-operated control valve (12), which allows continuous control with a signal ranging from 0~V to 10~V. This control signal regulates the operation of the globe valve, which moves linearly and is responsive to the pressure differential $\Delta p$ between the temperature control circuit and the circulation circuit.

External loads are connected to the interface $To6$ and $Ti6$, which can transfer heat $Q_L$ through the temperature control circuit. A check valve (31) prevents the incorrect supply of cooling water and ensures that heated water is directed back from the temperature control circuit to the circulation circuit. Furthermore, the system incorporates both permanently attached immersion sensors and contact sensors as temperature sensors. Each circuit is equipped with analogue pressure gauges to monitor pressure levels.

\subsubsection{Default Control Strategy}\label{sec:exp_1_3}

The default control strategy is a PI controller. Since most cooling systems involve inert mechanical components, the derivative component $K_D$ is not beneficial and hence, not used. The control behavior is influenced by specific conditions, which are manually defined. The impact of the non-constant circulation temperature is accounted for by calculating the difference between the set temperature and the actual circulation temperature. The proportional component $K_P$ is adjusted and varies inversely in proportion to this difference. Additionally, if the temperature difference becomes excessively high, the anti-wind-up adjusts the control output to either its maximum or minimum value.

\subsubsection{Eb-SbPG Setting}\label{sec:setting_sbpg}

We transform the controlled system into an EB-SbPG, as shown in Fig.~\ref{fig:closedloop_gt}, where it involves two players that are responsible for controlling the $K_P$ and $K_I$ values of the PID controllers associated with the motor-operated valve, see Component 12 in Fig.~\ref{fig:pid}. Notably, $K_D$ is not considered in the configuration of our test environment, as explained in Sec.~\ref{sec:exp_1_3}. Each player $i$ has two state information variables denoted as $s_i^1$ and $s_i^2$, which include (a) the absolute difference between the actual temperature of the temperature control circuit $T_T$ and the desired temperature of the temperature control circuit $T_{set}$, and (b) the absolute difference between the cooling water temperature in the circulation circuit $T_Z$ and the desired temperature of the temperature control circuit $T_{set}$. The states can be mathematically formulated as follows:
\begin{align}\label{eq:states1}
s_i^1 = \lvert T_T - T_{set}\lvert,
\end{align}
\begin{align}\label{eq:states2}
s_i^2 = \lvert T_Z - T_{set}\lvert.
\end{align}

\begin{figure}[t]
	\centering
	\includegraphics[width=1.0\linewidth,keepaspectratio]{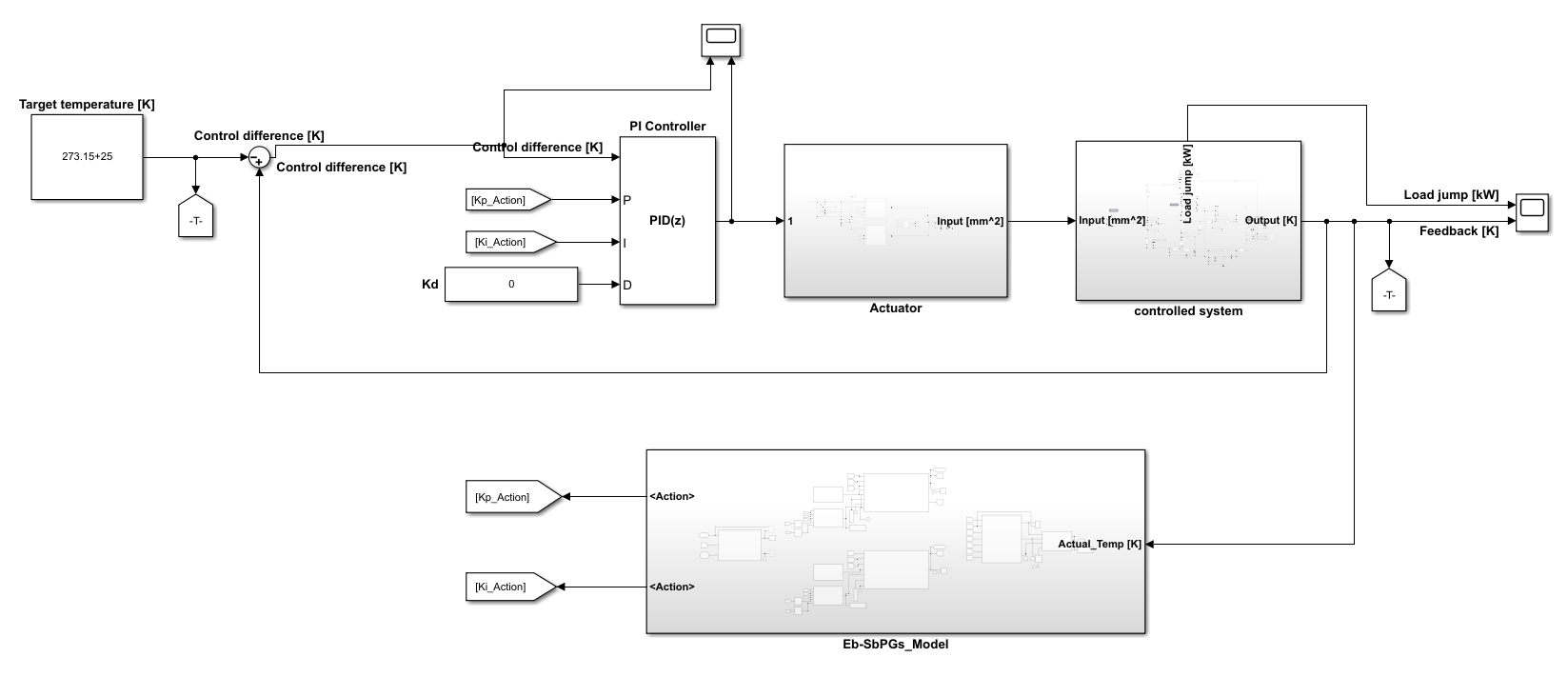}
	\caption{The Simulink model demonstrating the integration of the self-tuning PID controller and the closed-loop system.}
	\label{fig:simulink}
\end{figure}

The integration between the closed-loop control model and the Eb-SbPG's self-tuning mechanism is illustrated in Fig.~\ref{fig:simulink}. In this setup, the players adjust their actions, $K_P$ and $K_I$ values, in response to the actual system state whenever an event is triggered. Such an event is initiated when the actual temperature of the temperature control circuit $T_T$ falls outside the upper and lower threshold boundaries $\theta_h$ and $\theta_l$. This particularly occurs due to disturbances or set point changes of the temperature $T_{set}$. In our system, we aim to maintain $T_T$ within the range of $T_{set}\pm 0.5^{\circ}C$. As a result, $\theta_h$ and $\theta_l$ are set to $0.5^{\circ}C$ and $-0.5^{\circ}C$, respectively. During the training phase, after the end of an event, the players evaluate the outcomes of the selected actions during the event by measuring the utility value $U_i$ and subsequently update their performance maps accordingly.

\subsection{Results and Discussions}\label{sec:test2}

In this section, we present and analyze the experimental outcomes of the Eb-SbPGs for self-tuning PID controllers applied to the temperature control loop. First, we provide the results of the action set determination, as detailed in Sec.~\ref{sec:theory}. Subsequently, we examine both versions of utility functions, as outlined in Sec.~\ref{sec:exp_2_2}, and compare their performance against baselines with random constant $K_P$ and $K_I$ values. Finally, we verify the effectiveness of the proposed method by conducting a random load test within the system.

\subsubsection{Test Scenario}\label{sec:scenario}

To evaluate the system, we carry out multiple testing cycles, with each cycle lasting 2,000~sec and comprising a combination of 70\% high load utilization (4 kW) and 30\% low load utilization (2 kW) for the printing press machine. These load variations directly influence the heat generated by the printing press, which impacts the system's behavior. The system's behavior is also influenced by fluctuations in the cooling water temperature and by related components within the system, e.g. friction and heat from the pump.

\subsubsection{Determination of Action Set}\label{sec:res_a_set}

\begin{figure}[t]
	\centering
	\includegraphics[width=1.0\linewidth,keepaspectratio]{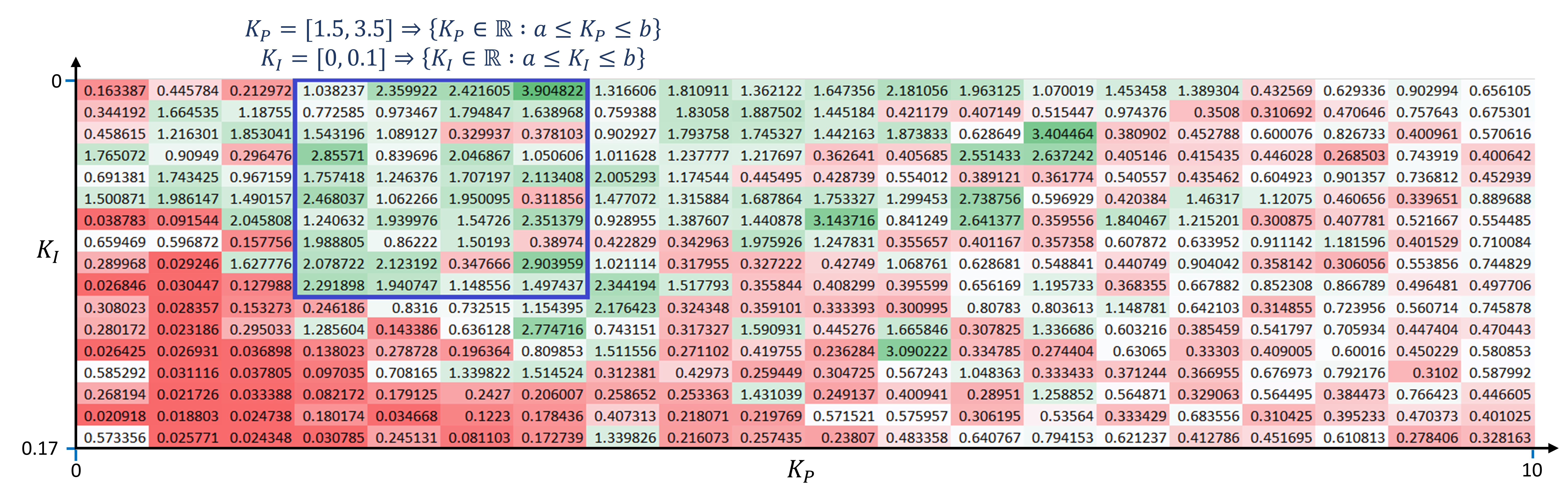}
	\caption{Results of automatic boundary detection of the action space.}
	\label{fig:boundary}
\end{figure}

We start by setting the boundaries of the action space for the $K_P$ and $K_I$ players to the ranges of $0-10$ and $0-0.17$, respectively. However, lengthy exploration time was required to reach optimal $K_P$ and $K_I$ values in different states due to the extensive action space. Consequently, to narrow down the action space and identify the potential optimal region of $K_P$ and $K_I$ values, we conduct a grid search procedure by discretizing the action space and assessing various $K_P$ and $K_I$ values using the utility function from Eq.~\eqref{eq:util_centralized_1} over a specific timeframe. Fig.~\ref{fig:boundary} shows the potentially optimal and stable region for the action spaces within $K_P \in [1.5, 3.5]$ and $K_I = [0, 0.1]$. Consequently, these action space boundaries are utilized throughout subsequent experiments.

\subsubsection{Baseline Control}\label{sec:baselines}

\begin{figure}[t]
	\centering
	\includegraphics[width=1.0\linewidth,keepaspectratio]{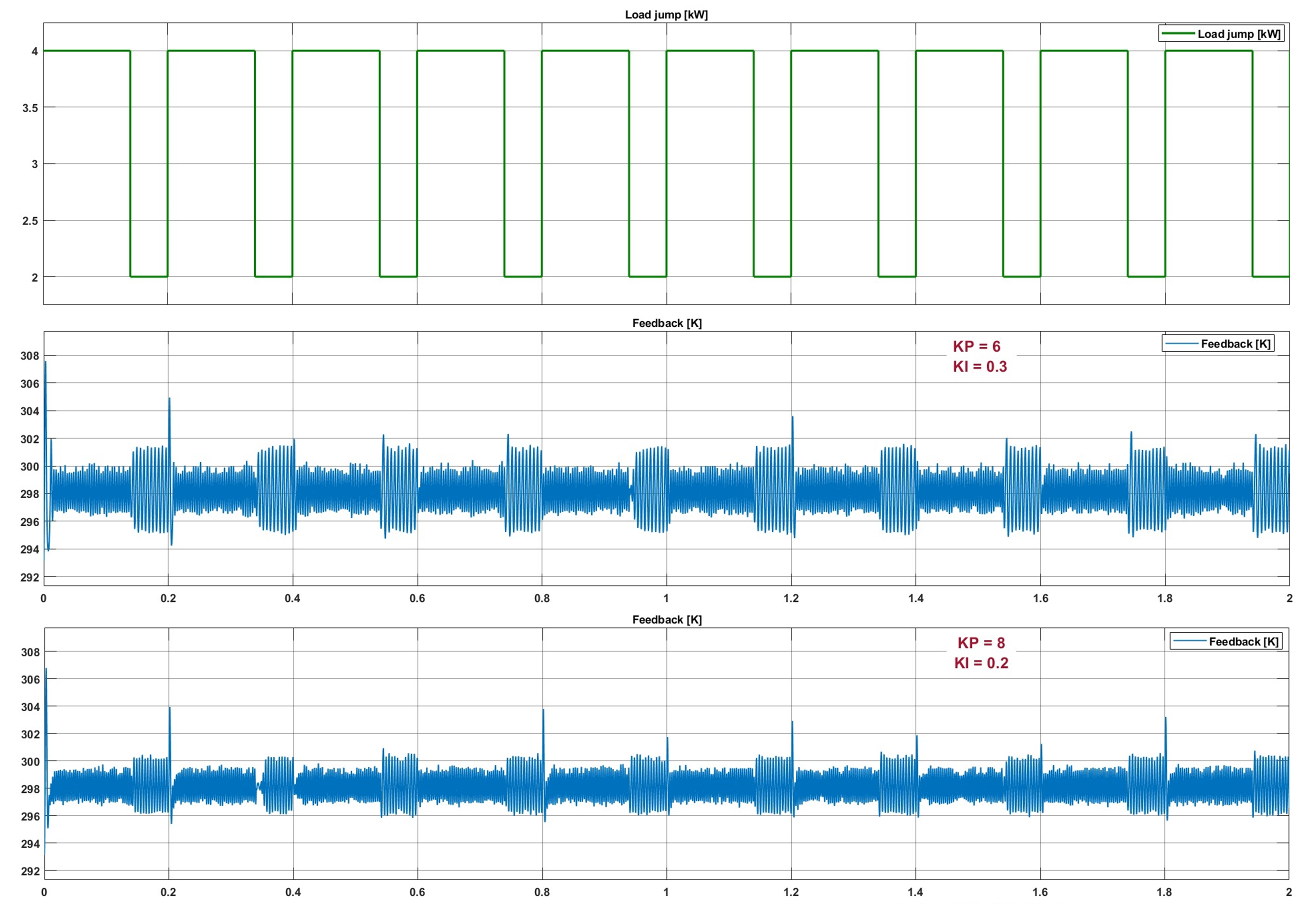}
	\caption{Testing results with a set of random constant $K_P$ and $K_I$ values.}
	\label{fig:res_rand_gains}
\end{figure}

As part of the baseline assessment, we conduct tests employing various $K_P$ and $K_I$ values without the integration of Eb-SbPGs, where the $K_P$ and $K_I$ values remain constant and not adaptable to system dynamics. This approach leads to challenges in determining the appropriate $K_P$ and $K_I$ values that would ensure system stability. The system exhibits pronounced oscillations and elevated overshoot, as depicted in Fig.~\ref{fig:res_rand_gains}.

\subsubsection{Results of Eb-SBPG}\label{sec:res_eb_sbpg}

\begin{figure}[t]
	\centering
	\includegraphics[width=1.0\linewidth,keepaspectratio]{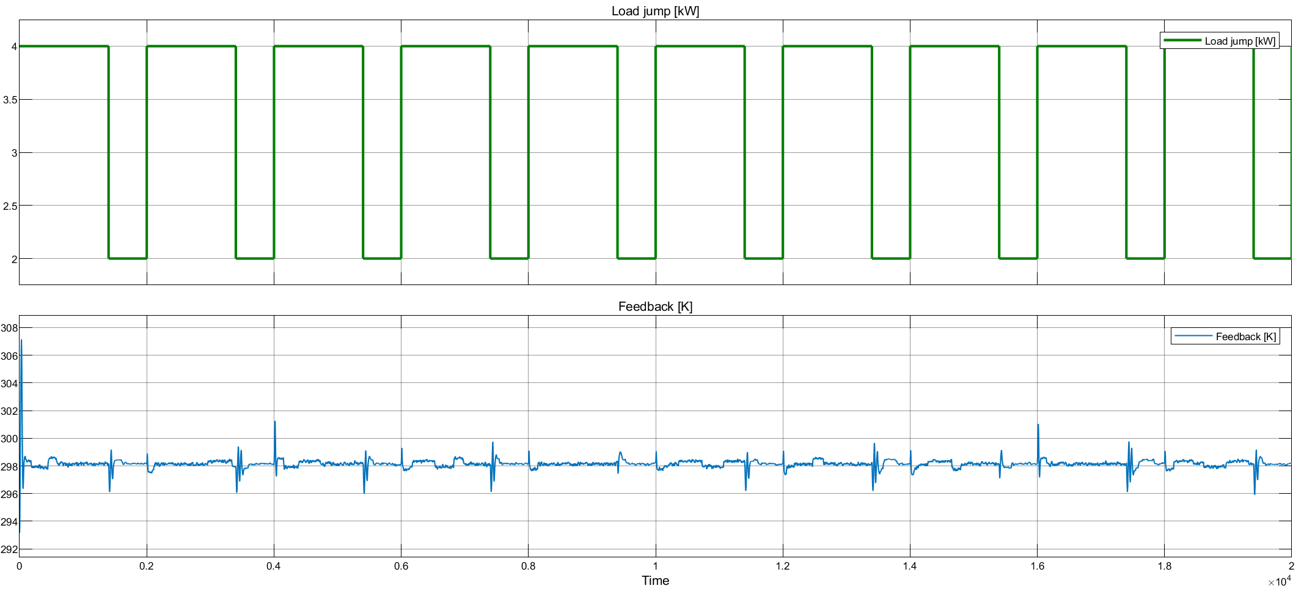}
	\caption{Testing results of Eb-SbPG with gradient-based learning using utility function type 1.}
	\label{fig:res_centralized1}
\end{figure}

\begin{figure}[t]
	\centering
	\includegraphics[width=1.0\linewidth,keepaspectratio]{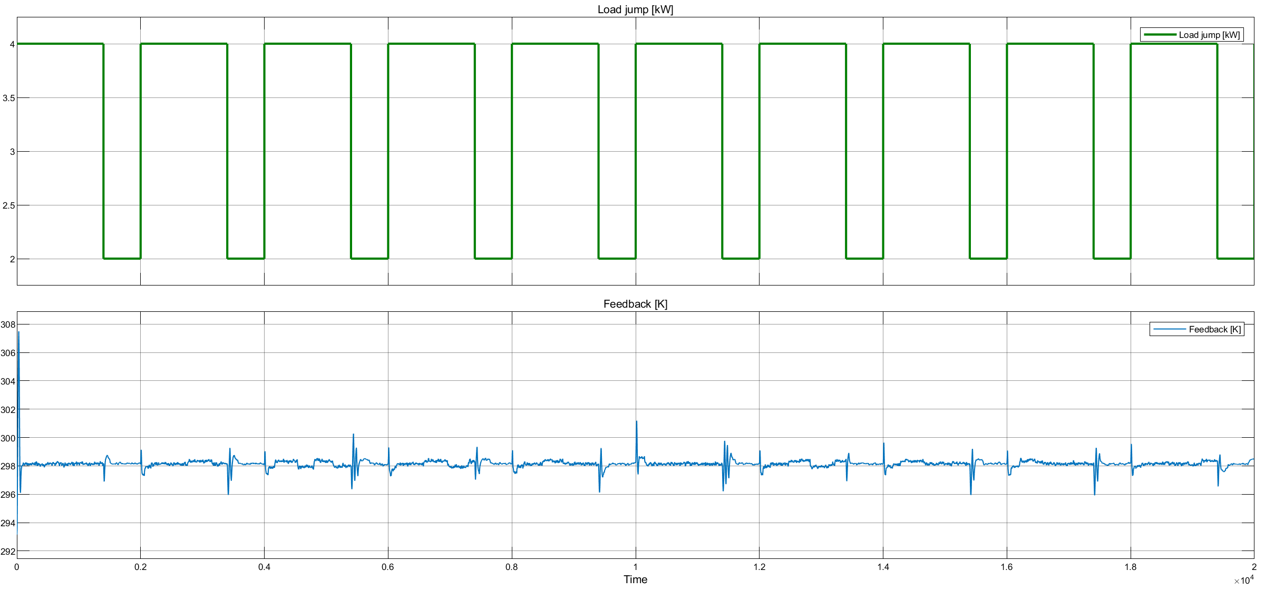}
	\caption{Testing results of Eb-SbPG with gradient-based learning using utility function type 2.}
	\label{fig:res_centralized2}
\end{figure}

We implement the Eb-SbPGs approach using utility function type 1 and type 2, with $\alpha_x$, $\alpha_y$, and $\alpha_i$ set to 0.3, 0.3, and 0.8 for player $K_P$, and 0.1, 0.1, and 10 for player $K_I$. The outcomes of the tuned controllers with both utility functions both exhibit stable and effective control, as shown in Fig.~\ref{fig:res_centralized1} and Fig.~\ref{fig:res_centralized2}. By comparing to the baseline, the effectiveness of the Eb-SbPG controller tuning is easily shown. This signifies that the players have effectively proposed optimal $K_P$ and $K_I$ values in both conditions and have demonstrated adaptability to the system's dynamics.

The testing results of the two utility functions display a similar response. It is challenging to observe visually which utility functions outperform the other. Therefore, we have compiled a performance evaluation table comparing both learning approaches in Table~\ref{tab:comp}. The table highlights that both learning methods are equally effective in reducing overshoot. However, utility function type 2 achieves a shorter settling time on average by approximately 4.5 seconds in comparison to utility function type 1.

\begin{table}[ht]
\renewcommand{\arraystretch}{1.3}
\caption{Performance comparisons between utility functions type 1 and 2.}
\label{tab:comp}
	\centering
\begin{tabular}{|c|c|c|c|}
\hline
\begin{tabular}[c]{@{}c@{}}Utility\\ Function\end{tabular} & \begin{tabular}[c]{@{}c@{}}Avg. Settling\\ Time (s)\end{tabular} & \begin{tabular}[c]{@{}c@{}}Maximum \\ Overshoot ($^{\circ}$K)\end{tabular} & \begin{tabular}[c]{@{}c@{}}Maximum\\ Undershoot ($^{\circ}$K)\end{tabular} \\ \hline\hline
Type 1 & 59.140 & 301.23 & 295.94 \\ \hline
Type 2 & 54.777 & 301.18 & 295.98 \\ \hline
\end{tabular}
\end{table}

\subsubsection{Active Action Sets Determination}\label{sec:ablations}

In the ablations, we analyze various components of our approach, particularly, the impact of active action set determination, a comparison of different learning algorithms for the Eb-SbPG, and we employ a random load test with varying temperature set points and load utilization. The set point temperature ranges from 295.15 K to 299.15 K, while the load utilization varies between 2 kW and 4 kW, with durations ranging from 600 to 1,600 seconds. Additionally, the cooling water temperature fluctuates between 292.15 K and 293.15 K, along with unmeasurable disturbances from system components.

\begin{figure}[t]
    \centering
	\includegraphics[width=1.0\linewidth,keepaspectratio]{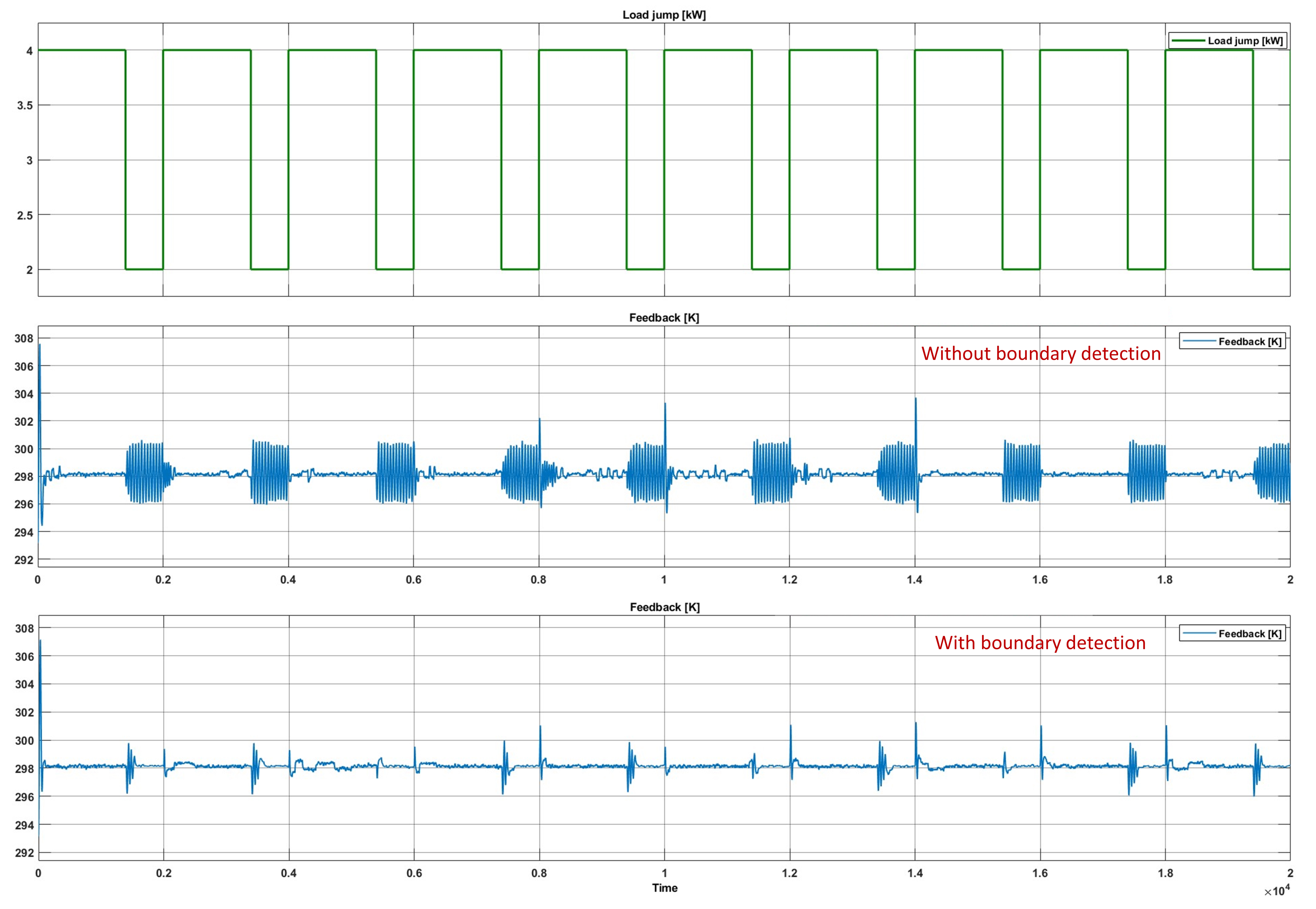}
	\caption{Testing results of centralized learning with and without automatic boundary detection.}
	\label{fig:activeactionset}
\end{figure}

\textbf{Active action sets determination}: We show the effect of actively determined action sets compared to a much wider action set in Fig.~\ref{fig:activeactionset}. Without automatic action set determination, it is evident that in high-load conditions, the system can be quickly stabilized and reach the desired set point. However, when the load conditions shift, the system encounters disturbances that lead to an oscillatory behavior. This suggests that the model fulfills its intended purpose, but it is not yet optimized. Meanwhile, with actively determined action sets, we observe that Eb-SbPGs' tuner facilitates rapid settling time to the set point in both high and low load conditions.

\begin{figure}[t]
	\centering
	\includegraphics[width=1.0\linewidth,keepaspectratio]{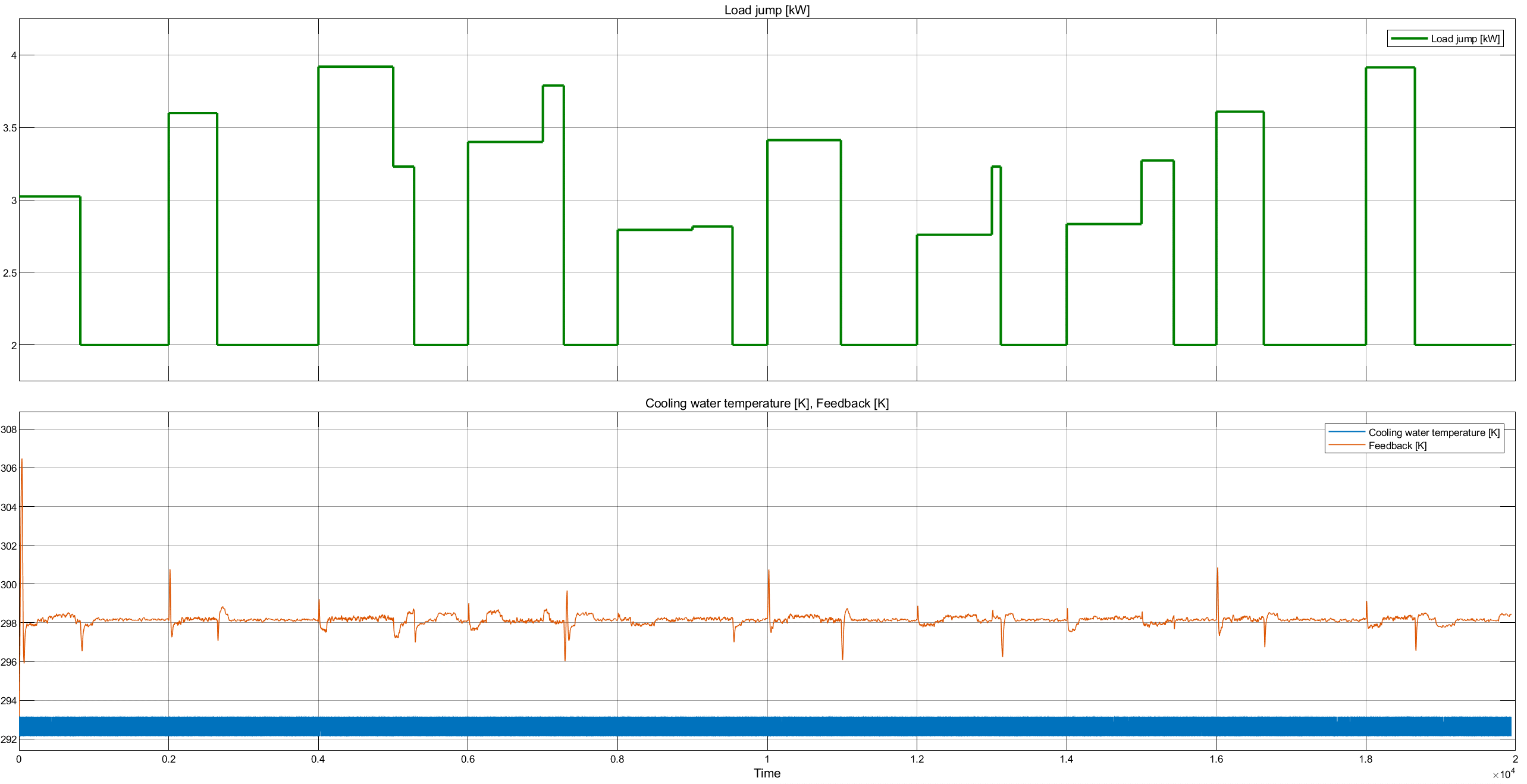}
	\caption{Validation results of Eb-SbPG with best response learning with variable load test and disturbances.}
	\label{fig:res_centralized_br}
\end{figure}

\textbf{Learning algorithms}: In this ablation, we compare two learning algorithms mentioned in Sec.~\ref{sec:learn_1} under static and random load tests with set point temperature of 298.15 K, i.e. gradients based learning~\cite{Yuwono2024} and best response learning~\cite{Schwung2022} as reported in Tab.~\ref{tab:complearn}. Additionally, Fig.~\ref{fig:res_centralized_br} presents the results of best response learning under the random load test, while the breakdown of gradient-based learning is provided in the next ablation study. As evident from the results, gradient based learning outperforms best response in settling time by a huge margin. Hence, we recommend using gradient based learning for PID controller tuning.
\begin{table}[ht]
\renewcommand{\arraystretch}{1.3}
\caption{Comparisons between EB-SbPG with gradient-based learning and best response learning.}
\label{tab:complearn}
	\centering
\begin{tabular}{|cccc|}
\hline
\multicolumn{1}{|c|}{\begin{tabular}[c]{@{}c@{}}Learning\\ Algorithm\end{tabular}} & \multicolumn{1}{c|}{\begin{tabular}[c]{@{}c@{}}Avg. Settling\\ Time\\ (s)\end{tabular}} & \multicolumn{1}{c|}{\begin{tabular}[c]{@{}c@{}}Maximum \\ Overshoot\\ ($^{\circ}$K)\end{tabular}} & \begin{tabular}[c]{@{}c@{}}Maximum\\ Undershoot\\ ($^{\circ}$K)\end{tabular} \\ \hline\hline
\multicolumn{4}{|c|}{Static Load Test} \\ \hline
\multicolumn{1}{|c|}{GB Learning} & \multicolumn{1}{c|}{54.777} & \multicolumn{1}{c|}{301.18} & 295.98 \\ \hline
\multicolumn{1}{|c|}{BR Learning} & \multicolumn{1}{c|}{67.425} & \multicolumn{1}{c|}{301.74} & 295.84 \\ \hline\hline
\multicolumn{4}{|c|}{Variable Load Test and Disturbances} \\ \hline
\multicolumn{1}{|c|}{GB Learning} & \multicolumn{1}{c|}{51.026} & \multicolumn{1}{c|}{299.85} & 296.12 \\ \hline
\multicolumn{1}{|c|}{BR Learning} & \multicolumn{1}{c|}{69.207} & \multicolumn{1}{c|}{300.91} & 295.97 \\ \hline
\end{tabular}
\end{table}

\begin{figure}[t]
	\centering
	\includegraphics[width=1.0\linewidth,keepaspectratio]{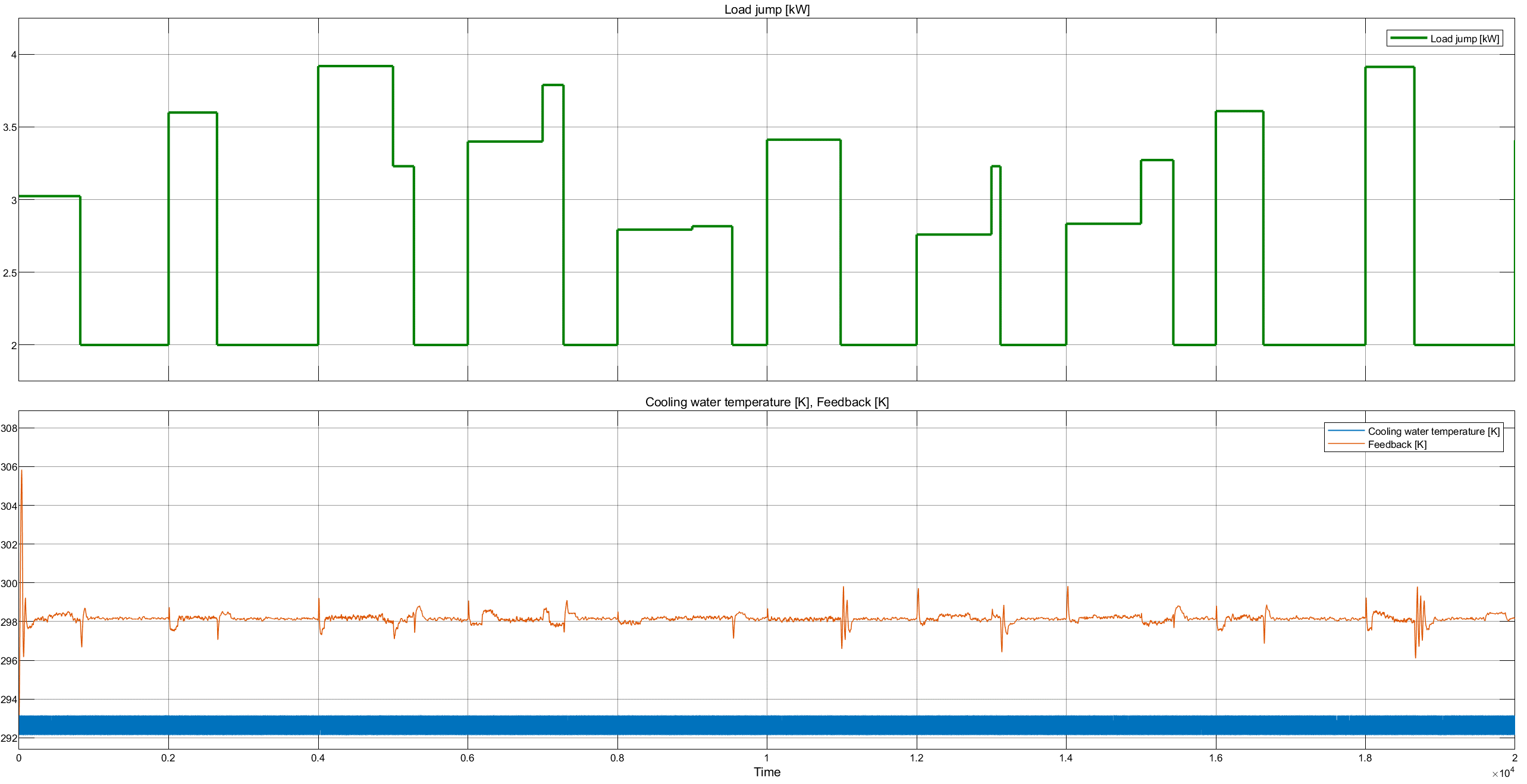}
	\caption{Validation results of Eb-SbPG with gradient-based learning with variable load test and disturbances.}
	\label{fig:res_centralized}
\end{figure}

\textbf{Random load tests}: In the actual printing machine system, the load is subject to fluctuations rather than remaining constant. It can change at any moment based on operational requirements. To validate our approach, we conduct a random load test using the proposed controller with utility function type 2. In this test, the sequence of high and low load conditions is arbitrary, which allows us to observe the results and determine the system's efficacy in unknown scenarios. Fig.~\ref{fig:res_centralized} illustrates that the model remains unaffected by the varying load cycles with the set point temperature of 298.15 K. The temperature quickly settles back to the desired set point temperature and the settling time in each scenario remains consistent. Moreover, we validate our approach under varying set point temperatures, which are summarized in Table~\ref{tab:comp_set}. This adaptability highlights the robustness of the Eb-SbPGs as self-tuning PID controllers.

\begin{table}[b]
\renewcommand{\arraystretch}{1.3}
\caption{Validation results of Eb-SbPG under varying set point temperatures and load utilization.}
\label{tab:comp_set}
	\centering
\begin{tabular}{|c|c|c|c|}
\hline
\begin{tabular}[c]{@{}c@{}}Set Point\\ Temperature \\($^{\circ}$K)\end{tabular} & \begin{tabular}[c]{@{}c@{}}Avg. Settling\\ Time \\(s)\end{tabular} & \begin{tabular}[c]{@{}c@{}}Maximum \\ Overshoot \\($^{\circ}$K)\end{tabular} & \begin{tabular}[c]{@{}c@{}}Maximum\\ Undershoot \\($^{\circ}$K)\end{tabular} \\ \hline\hline
295.15 & 73.422 & 297.09 & 293.90 \\ \hline
296.15 & 64.575 & 297.94 & 294.94 \\ \hline
297.15 & 58.998 & 299.87 & 295.40 \\ \hline
298.15 & 51.026 & 299.85 & 296.12 \\ \hline
299.15 & 69.436 & 301.96 & 296.65 \\ \hline
\end{tabular}
\end{table}

\section{Conclusions}\label{sec:conclusion}

We presented a novel approach for PID-controller parameter tuning by extending SbPGs to EB-SbPGs, introducing a novel event-based game structure where players only update their utility once an event is triggered due to disturbances or set-point changes. We have analyzed convergence and stability properties of the resulting game, assuring convergence within a stable parameter set under mild assumptions. We have exploited the potential of EB-SbPGs to construct an intelligent self-tuning PID controller using a model-free approach and validated it on the temperature control loop of a printing press machine. The results of our proposed intelligent PID controller show significant improvements over the default control strategy, where the controller demonstrates its adaptability by dynamically adjusting the $K_P$, $K_I$, and $K_D$ values in response to the actual system states.

In future research, we plan to explore alternative game structures to further enhance the tuning capabilities of the approach. Also, we want to employ individual, per player utility functions such that specific performance metrics are handled by specific players. Another future work will address the direct replacement of the PID controller with an Eb-SbPGs controller, where the players directly propose control signals instead of adjusting the parameter values. 

\bibliographystyle{IEEEtran}
\bibliography{sample}

\end{document}